\documentclass[letterpaper, 10 pt, conference]{ieeeconf} 
\IEEEoverridecommandlockouts 
\usepackage{graphicx}
\usepackage{amsmath,amssymb,amsfonts}
\usepackage{cite}
\usepackage[breaklinks,hidelinks,draft]{hyperref}
\usepackage{bbm}
\usepackage{subcaption}
\usepackage{censor}

\title{\LARGE \bf \textbf{Towards an Adaptive Social Game-Playing Robot: An Offline Reinforcement Learning-Based Framework}}

\author{
    Soon Jynn Chu$^{1}$, Raju Gottumukkala$^{1}$, Alan Barhorst$^{1}$%
    \thanks{$^{1}$ Department of Mechanical Engineering, University of Louisiana at Lafayette, Lafayette, LA 70504, USA.
    \texttt{\{soonjynn.chu1, alan.barhorst, raju.gottumukkala\}@louisiana.edu}}%
}

\begin{document}

\maketitle
\thispagestyle{empty}
\pagestyle{empty}

\begin{abstract}

HRI research increasingly demands robots that go beyond task execution to respond meaningfully to user emotions. This is especially needed when supporting students with learning difficulties in game-based learning scenarios. Here, the objective of these robots is to train users with game-playing skills, and this requires robots to get input about users' interests and engagement. In this paper, we present a system for an adaptive social game-playing robot. However, creating such an agent through online RL requires extensive real-world training data and potentially be uncomfortable for users. To address this, we investigate offline RL as a safe and efficient alternative. We introduce a system architecture that integrates multimodal emotion recognition and adaptive robotic responses. We also evaluate the performance of various offline RL algorithms using a dataset collected from a real-world human-robot game-playing scenario. Our results indicate that BCQ and DDQN offer the greatest robustness to hyperparameter variations, whereas CQL is the most effective at mitigating overestimation bias. Through this research, we aim to inform the selection and design of reliable offline RL policies for real-world social robotics. Ultimately, this work provides a foundational step toward creating socially intelligent agents that can learn complex and emotion-adaptive behaviors entirely from offline datasets, ensuring both human comfort and practical scalability.
\end{abstract}

\section{Introduction}
Social robotics is a subset of human-robot interaction (HRI) that focuses on designing robots capable of engaging with users through social behaviors, communication, and interaction patterns. They are increasingly being developed as conversational agents, educational partners, and personal assistants. For example, the PARO robot boosts happiness and reduces pain perception in older adults \cite{geva_touching_2020}, while the Pepper robot encourages activity in healthcare settings \cite{blindheim_promoting_2023}. The ability to perceive and respond to the user's emotions presents opportunities for deeper engagement, but the relationship between emotional responsiveness and user trust remains complex and context-dependent. Achieving this requires both accurate emotion recognition and adaptive behavior selection. However, naive mappings from detected emotions to robot actions often produce responses that feel unnatural or inappropriate. 

While Wizard-of-Oz methods \cite{marge_applying_2017} were widely used to prototype social behaviors in earlier works, this approach is not scalable for real-world deployment. The need for a human operator to constantly control the robot highlights the challenge of achieving effective and autonomous HRI. 


Reinforcement learning (RL) offers a data-driven approach to learning nuanced emotional response strategies that adapt to diverse user behaviors. Robots learn optimal emotion response policies through trial-and-error interaction with users \cite{sutton_reinforcement_2018}. Online RL learns through direct interaction with users, requiring numerous trial-and-error episodes to converge to optimal policies. In HRI contexts, this approach becomes prohibitively expensive due to the need for human participants, lengthy interaction sessions, and potential risk of unsafe and socially inappropriate behaviors that might be harmful or cause discomfort for participants. 

Offline RL has gained attention as a practical alternative to online RL. It has been used in various applications such as controlling navigation for omni-wheeled robots \cite{amarasiri_investigating_2024} and developing effective and safer insulin dosing policies for blood glucose control \cite{emerson_offline_2023}. Unlike online RL, which depends on extensive real-time exploration, offline RL enables policy learning entirely from pre-collected datasets. This makes it well-suited for HRI, where arbitrary exploration can be unsafe, costly, or risk undermining user trust. By leveraging logged interaction data, offline RL enables robots to develop adaptive response policies while avoiding the inefficiencies and trial-and-error risks associated with user interactions. Recent studies (e.g., \cite{hong_multimodal_2020, tielman_adaptive_2014, khabbaz_adaptive_2019, ahmad_emotion_2022}) have explored emotion-aware HRI; however, the field lacks a comprehensive framework for systematically developing and evaluating offline RL policies tailored to emotionally adaptive social robots. 


In this paper, we present a novel system for an adaptive social game-playing robot using an offline RL approach. We aim to establish a baseline for developing safe, adaptive social robots that learns from pre-collected data without risking poor user experiences during training. Our primary contributions are:

\begin{itemize}
   \item A multimodal offline RL framework that converts complex user signals into an RL state-space. We used a game-playing scenario as a structured yet socially rich testbed that elicits diverse emotions while posing minimal risk.
    
    \item A comparative study determining which offline RL algorithms are the most robust to hyperparameter choices and the least prone to overestimation or underestimation under data-scarce conditions. The findings inform the selection of stable and reliable algorithms for future social robotics settings.
\end{itemize} 

We begin in Section \ref{sec:related_works} by reviewing the related literature on social robotics and offline RL. Sections \ref{sec:system_arch} and \ref{sec:methodology} outline our system architecture and methodology, including the offline RL formulation, data collection, implementation, and evaluation protocol. Our experimental results and analysis are presented in Section \ref{sec:results}, followed by a conclusion in Section \ref{sec:conclusion}.

\section{Related Works}
\label{sec:related_works}

\subsection{Adaptive Response Generation using RL}
Earlier works on emotion-adaptive responses primarily relied on rule-based approaches. For example, \cite{liu_eeg-based_2013} developed an EEG-based system that recognized users' emotional states in terms of valence, arousal, and dominance, and adapted advertisements accordingly through predefined rules based on the detected emotions. While effective in constrained settings, such rule-based systems lack flexibility and struggle to generalize to the variability and unpredictability of real-world user behavior.

RL has become a well-known approach for automated mapping between recognized states and actions, and it has been extensively researched for response generation in HRI. For example, \cite{leite_modelling_2011} developed a chess companion robot that used visual cues and game context to model children’s states and applied multi-armed bandits to generate empathetic responses tailored to each child. Similarly, \cite{park_modelfree_2019} designed a learning companion robot that leveraged verbal and non-verbal cues from children and trained a Q-learning model to adapt stories and interactions based on each learner’s engagement and progress. While these works demonstrate the potential of RL in adaptive HRI, they rely on online training, which is time- and resource-intensive, making algorithmic convergence investigations and cross-algorithm comparisons difficult.


\subsection{Offline RL in HRI}
Social robotics has yet to fully leverage offline RL methodologies, despite their clear advantages for safe, efficient, and scalable HRI training. By learning policies entirely from pre-collected datasets, offline RL eliminates the need for continuous human supervision and unsafe online exploration, making it a promising pathway for developing autonomous and scalable systems. For instance, \cite{hussain_training_2022} employed a Sequential Random Deep Q-Network, an offline RL algorithm, to train a robot head for generating backchanneling behaviors in a storytelling game. Similarly, \cite{qi_adaptive_2023} proposed a multimodal adaptive RL framework in a rock–paper–scissors interaction, combining Q-learning and Sarsa with ensemble classifiers to improve robustness against noisy inputs. While these works demonstrate the feasibility of offline approaches, their reliance on narrow input modalities, such as audio cues \cite{hussain_training_2022} or hand-motion signals \cite{qi_adaptive_2023}, limits the scope of affective understanding. Without incorporating internal affective states, robots may fail to fully capture user emotions and thus struggle to generate truly adaptive and socially meaningful responses.


\section{System Architecture Overview}
\label{sec:system_arch}
\begin{figure*}[htbp]
    \centering
    \includegraphics[width=0.75\textwidth]{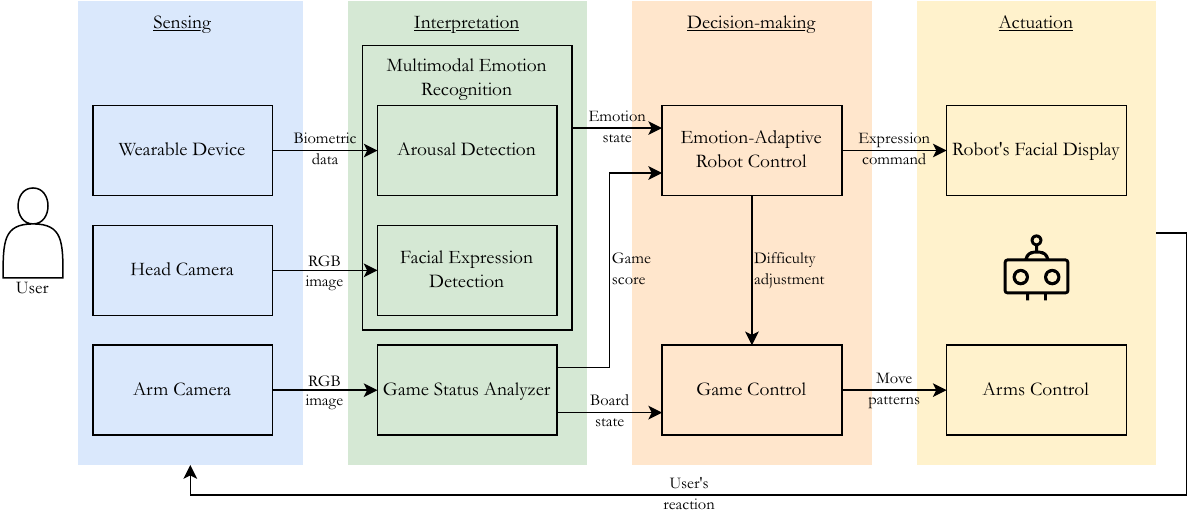}
    \caption{System Architecture}
    \label{fig:system_architecture}
\end{figure*}

Figure \ref{fig:system_architecture}  provides an overview of our four-layer system architecture, which processes multimodal user data to generate adaptive robot responses. The four layers -- sensing, interpretation, decision-making, and actuation -- work in tandem to enable the robot to perceive, understand, and react to both the user's emotional state and the evolving game context in real time.

The sensing layer captures both user and game states through three hardware components: a wearable device (Empatica E4), a head-mounted camera, and an arm-mounted camera. The wearable device records biometric signals, providing continuous physiological data that reflects the user's internal arousal level. The head-mounted camera captures RGB images of the user's face, enabling the system to analyse facial expressions as indicators of emotional state. The arm-mounted camera monitors the game board, providing a visual record of piece positions and game progression.

The interpretation layer converts raw sensor data into meaningful information for the robot's decision-making. This is achieved through two primary modules: the multimodal emotion recognition module and the game status analyzer. The emotion recognition module processes biometric signals from the sensing layer to detect physiological arousal, while also using head camera images to classify the user's facial expressions into discrete emotional categories. Simultaneously, the game status analyzer extracts the current board state from the arm camera feed and computes a game score based on the positions and relative value of remaining pieces. Together, these modules provide the robot with a holistic understanding of both the user's emotional condition and the ongoing game context.

Within the decision-making layer, two modules handle distinct but interconnected functions: emotion-adaptive robot control and game control. The emotion-adaptive robot control module receives emotion state data and game scores from the interpretation layer, using this information to generate contextually appropriate adaptive responses. An expression command is produced and forwarded to the actuation layer for display on the robot's screen, while a difficulty adjustment signal is simultaneously sent to the game control module. The game control module uses this adjustment alongside the current board state to calibrate the difficulty of gameplay and determine suitable move patterns, which are then routed to the arms control module for physical execution.

The actuation layer constitutes the robot's physical output interface, comprising the robot's facial display and the arms control module. The facial display renders expressive messages to communicate empathetically with the user. The arms control module executes the move patterns determined by the game control module, physically manipulating the game pieces on the board. The user reacts to the robot's actions, which is captured by the sensing layer, closing the system loop and ensuring that the robot continuously adapts to the evolving state of both the user and the game.

\section{Methodology}
\label{sec:methodology}
\subsection{Reinforcement Learning Setup}
To develop our emotion-adaptive robot control module, we can formulate this as an Markov Decision Process, which is represented by the tuple ($s_t$, $a_t$, $r_t$, $s_{t+1}$), where $s_t$ and $s_{t+1}$ are the current and next state of the user at timestep $t$, $a_t$ is the action of the robot after observing $s_t$, and $r_t$ is the reward for the robot for taking $a_t$ at $s_t$. 

We define $s_t$ based on the user and game status as a function of the game state (\textit{GS}), facial emotion (\textit{FE}), and physiological arousal (\textit{PA}). \textit{GS} represents the game status, indicating if the user is winning, losing, or drawing. \textit{FE} represents the user's facial emotion, indicating if the user is expressing an angry, happy, or neutral facial expression. \textit{PA} represents the user's physiological arousal state, indicated as absent or present. Conversely, $a_t$ is defined based on the robot's action as a function of facial display (\textit{FD}) and difficulty adjustment (\textit{DA}). \textit{FD} is the facial expression that the robot wants to convey to the user, where the robot selects from angry, happy, or neutral. \textit{DA} is how the robot seeks to adjust the game level, which can be harder, easier, or constant. In total, there are 18 states and 9 actions in this RL setup. 


Our reward function is designed to balance the robot's competitive performance with the user's emotional state, as defined in Equation \ref{eqn:reward_func}. The total reward consists of two components: $GameScore$ and $EmoScore$, which are bounded to the interval of [-1,1]. The $GameScore$ is defined as +1.0 when the user is losing and -1.0 when the user is winning, encouraging the robot to maintain an appropriate level of challenge. The $EmoScore$, defined in Equation \ref{eqn:emo_score}, captures the user's affective response. Since EDA reflects the intensity of emotional arousal, we use the number of skin conductance response (SCR) peaks as a measure of emotional intensity \cite{braithwaite_guide_2013}. This intensity is modulated by $FEScore$, which encodes +1.0 for positive expressions (e.g., happiness) and -1.0 for negative expressions (e.g., anger). We use the exponential saturation function $1-e^{-x}$ to map SCR peak counts to a bounded emotional intensity score, ensuring outputs are constrained to [0,1]. 

\begin{equation}
    \label{eqn:reward_func}
    r_t = \frac{1}{3} \Bigl[GameScore_{t+1} + 2 \times EmoScore_{t+1}\Bigr]
\end{equation}

\begin{equation}
    \label{eqn:emo_score}
    EmoScore = FEScore \times (1 - e^{-\#SCRPeaks})
\end{equation}

\subsection{Data Collection} 
We collected our dataset in the form of $D = \{ \{ s_t, a_t, r_t, s_{t+1} \}_{t=0} ^ {T-1} \}_{i=0}^{N-1}$ by logging the states and decisions of the robot, where $T$ is the termination timestep, and $N$ is the number of episodes. We have chosen American checkers as the game to be played, as it has a smaller learning curve. To ensure fairness, the robot takes an action sampled from a random uniform distribution for each episode. The participants choose the first player. No personally identifiable features are collected in the study. Moreover, consent to use the data for this work was obtained from each subject before and after the data collection. 

\begin{figure}[htbp]
    \centering
    \includegraphics[width=0.475\textwidth]{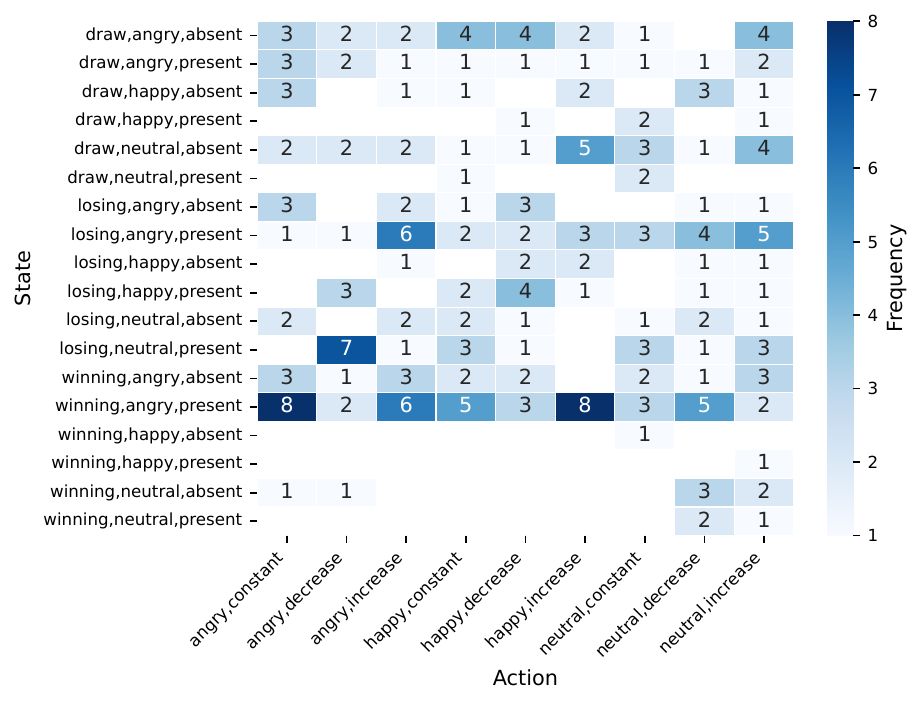}
    \caption{Frequency of Each State-Action Pair}
    \label{fig:state_action_pairs}
\end{figure}

The dataset was gathered from five participants. A total of 232 steps is collected, with 46.4$\pm$11.6 steps for each participant. The experiment duration for each participant is 36.1$\pm$7.0 minutes. The visitation frequency of each state-action pair is shown in Figure \ref{fig:state_action_pairs}. Overall, the dataset exhibits partial state-action coverage, with a coverage of 63\%, reflecting the practical constraints and limited coverage typical of real-world HRI data collection.

\begin{figure}[htbp]
    \centering
    \includegraphics[width=0.45\textwidth]{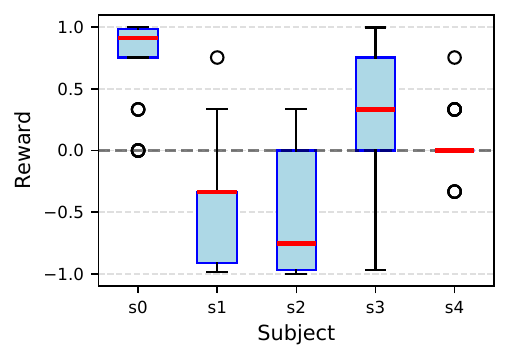}
    \caption{Reward Distribution for Each Subject}
    \label{fig:dataset_reward}
\end{figure}

Figure \ref{fig:dataset_reward} illustrates the reward distribution of each subject. The total distribution is somewhat balanced, with 38\% in the positive, 32\% in the negative, and 30\% in the zero region. Although the dataset is underexplored and does not represent an expert policy, it remains valuable for offline RL, as it can benefit from diverse and even imperfect data, enabling the agent to learn from both favorable and unfavorable outcomes. 


\subsection{Implementation Details}
The system architecture was implemented on the Baxter robot using ROS 1. Facial emotion recognition was performed using a mini-Xception network trained on the FER2013 dataset \cite{arriaga_realtime_2017}. The model was trained using the dataset filtered into three emotion classes: angry, happy, and neutral. The implementation protocols were otherwise kept consistent with those described in the original paper. For checkers piece detection, we collected and annotated a custom dataset, augmented it using RoboFlow, and trained a YOLOv8 model using the default hyperparameters \cite{jocher_yolo_2023}. The game score was calculated as the difference in the number of checker pieces between the user and the robot, as the goal of the game is to remove all of the opponent's pieces from the board.

For the emotion-adaptive RL module, we experimented on six offline RL algorithms, which are: Neural Fitted Q Iteration (NFQ) \cite{riedmiller_neural_2005}, Deep Q-Network (DQN) \cite{mnih_humanlevel_2015}, Double DQN (DDQN) \cite{vanhasselt_deep_2016}, Soft-Actor Critic (SAC) \cite{christodoulou_soft_2019}, Batch Constrained Q-Learning (BCQ) \cite{fujimoto_benchmarking_2019}, and Conservative Q-Learning (CQL) \cite{kumar_conservative_2020}.



We utilized the implementation in the \verb|d3rlpy| \cite{seno_d3rlpy_2022} library to train our RL agents. Since offline RL can be sensitive to hyperparameters, we conducted hyperparameter tuning for each algorithm \cite{paine_hyperparameter_2020}. The hyperparameters we chose to tune were referenced from \cite{paine_hyperparameter_2020} and their associated values are listed in Table \ref{tab:hyperparameters}. The values are selected based on common practices in typical neural network implementations, but are adjusted to better suit the constraints of small datasets. 

\begin{table}[htbp]
    \centering
    \caption{Selected Hyperparameters and Value Choices}
    \label{tab:hyperparameters}
    \begin{tabular}{ll}
    \hline
    \multicolumn{1}{c}{Hyperparameter} & \multicolumn{1}{c}{Value} \\ \hline
    Learning rate                        & 0.01, 0.001             \\
    Batch size                           & 8, 16                   \\
    Hidden layer                         & 1, 2                    \\
    Hidden units                         & 8, 16, 32               \\
    Activation                           & ReLU, Tanh              \\
    \hline
    \end{tabular}
\end{table}


We performed a grid search to find the optimal hyperparameter settings for each algorithm, resulting in 288 algorithm-hyperparameter pairs. The algorithms are trained for 10k steps ($\sim$ 50 epochs) and evaluated every 200 steps ($\sim$ 1 epoch). We chose the Adam optimizer as it has shown to converge faster and often requires no tuning parameters. Other hyperparameters are kept at their default values. Each state is one-hot encoded into a list of 8 elements, with the first three slots for $GS$, the subsequent three for $FE$, and the last slot for $PA$.

We employed Stratified Group K-Fold with 3 splits during training as a cross-validation technique. This is to provide enough coverage of positive and negative signals in both the train and test sets and facilitate generalization on unseen subjects, which is important in HRI and data-scarce settings where user-specific physiological and behavioral patterns can vary significantly. Overall, 864 training runs were conducted. 

\subsection{Evaluation Method}
Unlike online RL, agents in offline RL have no access to the environment and therefore cannot interact to obtain cumulative rewards. For discrete state-action spaces, we evaluate policies by estimating the expected value from the initial state distribution, defined as $\hat{V}(s_0) = \mathbb{E}_{s_0 \sim D} [Q(s_0, a_0)]$ \cite{paine_hyperparameter_2020}. The intuition is that we want the model to estimate state-action values that are ``conservative" enough such that they do not excessively overestimate or underestimate, even if they are out of distribution (OOD). The ground truth from the dataset for each episode can be estimated as $\sum_{t=0}^{T-1} \gamma^t r$, where $\gamma$ is the discount factor, which shares the same value as the training hyperparameter with a value of 0.99.


\section{Results \& Discussion}
\label{sec:results}

\subsection{Hyperparameter Sensitivity}
\begin{figure}[htbp]
    \centering
    \includegraphics[width=0.45\textwidth]{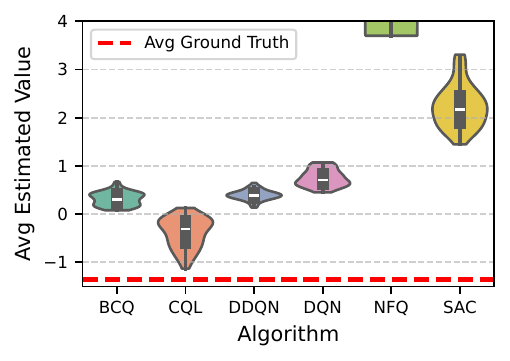}
    \caption{$\hat{V}(s_0)$ Distribution for Each Algorithm}
    \label{fig:algo_value_dist}
\end{figure}

This section evaluates the sensitivity of each algorithm to varying hyperparameters. The performance of a run is taken at the step with the lowest squared error relative to the ground truth. The average $\hat{V}(s_0)$ is then compared against the average ground truth value of -1.36 across each fold. Table \ref{tab:val_stats} summarizes these statistics. DDQN and BCQ show the lowest standard deviation (STD) at 0.10 and 0.14, respectively. Conversely, NFQ shows the highest variance, with an STD of 51.47 and a maximum $\hat{V}(s_0)$ reaching 323.68. The distribution of $\hat{V}(s_0)$ for each algorithm, shown in Figure \ref{fig:algo_value_dist}, shows a general trend across all algorithms to overestimate $\hat{V}(s_0)$. Notably, CQL produced the predictions that clustered closest to the ground truth line compared to the other algorithms. 

The low variance observed in DDQN and BCQ indicates that these algorithms are highly robust to hyperparameter changes, making them reliable choices for deployments where extensive hyperparameter tuning is computationally restricted. On the other hand, NFQ's extreme STD and high $\hat{V}(s_0)$ overestimation point to severe instability during learning, making it highly sensitive. CQL is the most effective in mitigating the overestimation bias by conservatively lower-bounding the value estimates, leading to representations that align much more closely with the true environment dynamics \cite{kumar_conservative_2020}.

\begin{table}[htbp]
\centering
\caption{Summary Statistics of the Average $\hat{V}(s_0)$ for Each Algorithm}
\label{tab:val_stats}
\begin{tabular}{lrrrr}
\hline
\multicolumn{1}{c}{Algorithm} & \multicolumn{1}{c}{Mean} & \multicolumn{1}{c}{STD} & \multicolumn{1}{c}{Min} & \multicolumn{1}{c}{Max} \\ \hline
BCQ  & 0.32  & 0.14  & 0.08  & 0.67   \\
CQL  & -0.39 & 0.31  & -1.13 & 0.13   \\
DDQN & 0.39  & 0.10  & 0.14  & 0.65   \\
DQN  & 0.75  & 0.17  & 0.45  & 1.07   \\
NFQ  & 29.75 & 51.47 & 3.70  & 323.68 \\
SAC  & 2.24  & 0.44  & 1.45  & 3.31   \\ \hline
\end{tabular}
\end{table}

\subsection{Algorithm Benchmarking and Optimal Configuration}
Table \ref{tab:algo_performance} shows the lowest absolute error achieved by each algorithm along with their best hyperparameter configurations. CQL achieved the best performance, with the lowest error at 0.23. This is followed by BCQ and DDQN that yield similar performance at 1.44 and 1.50 errors, respectively. NFQ exhibits the highest error at 5.06. Regarding best hyperparameters, there was no uniform preference across algorithms for learning rate, batch size, or network architecture (hidden layers and units). However, the ReLU activation function consistently emerged as the best choice across all algorithms, despite the state and reward spaces being strictly constrained to [-1, 1], which is more appropriate for the Tanh activation, as it outputs in this range.

\begin{table}[htbp]
\centering
\caption{Best Performance (Lowest Absolute Error) for Each Algorithm Along with their Hyperparameter Settings}
\label{tab:algo_performance}
\resizebox{\columnwidth}{!}{%
\begin{tabular}{lrrrrrr}
\hline
\multicolumn{1}{c}{Algorithm} &
  \multicolumn{1}{c}{\begin{tabular}[c]{@{}c@{}}Learning\\ Rate\end{tabular}} &
  \multicolumn{1}{c}{\begin{tabular}[c]{@{}c@{}}Batch\\ Size\end{tabular}} &
  \multicolumn{1}{c}{\begin{tabular}[c]{@{}c@{}}Hidden\\ Layer\end{tabular}} &
  \multicolumn{1}{c}{\begin{tabular}[c]{@{}c@{}}Hidden\\ Unit\end{tabular}} &
  \multicolumn{1}{c}{Activation} &
  \multicolumn{1}{c}{Error} \\ \hline
BCQ  & 0.001 & 8  & 2 & 8  & ReLU & 1.44 \\
CQL  & 0.01  & 16 & 1 & 32 & ReLU & 0.23 \\
DDQN & 0.001 & 8  & 2 & 8  & ReLU & 1.50 \\
DQN  & 0.01  & 16 & 1 & 16 & ReLU & 1.81 \\
NFQ  & 0.001 & 8  & 2 & 32 & ReLU & 5.06 \\
SAC  & 0.01  & 16 & 2 & 16 & ReLU & 2.80 \\ \hline
\end{tabular}%
}
\end{table}

\subsection{Convergence and Stability}
\begin{figure*}
    \centering
    
    \begin{subfigure}{0.3\textwidth}
        \includegraphics[width=\linewidth]{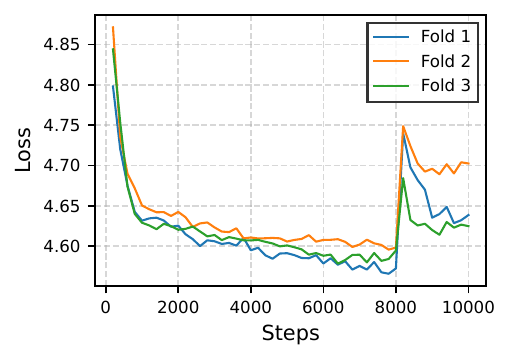}
        \caption{BCQ}
    \end{subfigure}
    \hfill
    \begin{subfigure}{0.3\textwidth}
        \includegraphics[width=\linewidth]{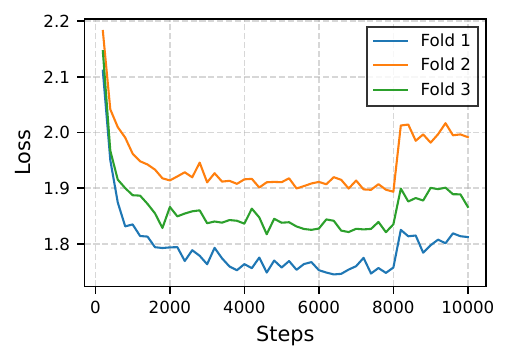}
        \caption{CQL}
    \end{subfigure}
    \hfill
    \begin{subfigure}{0.3\textwidth}
        \includegraphics[width=\linewidth]{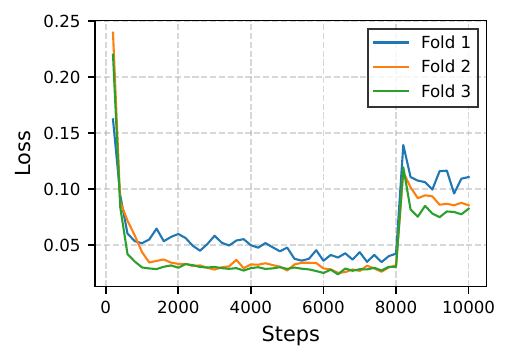}
        \caption{DDQN}
    \end{subfigure}
    
    \begin{subfigure}{0.3\textwidth}
        \includegraphics[width=\linewidth]{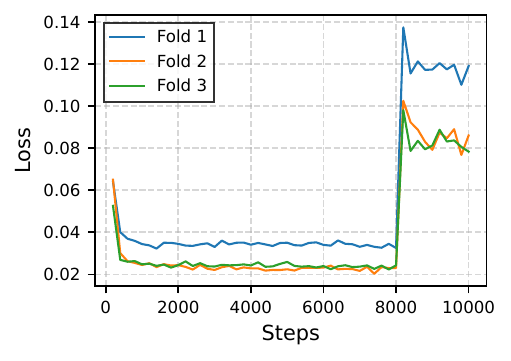}
        \caption{DQN}
    \end{subfigure}
    \hfill
    \begin{subfigure}{0.3\textwidth}
        \includegraphics[width=\linewidth]{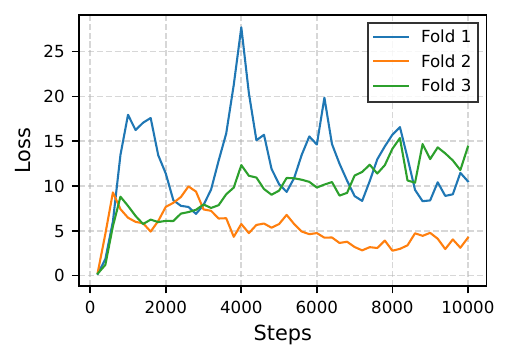}
        \caption{NFQ}
    \end{subfigure}
    \hfill
    \begin{subfigure}{0.3\textwidth}
        \includegraphics[width=\linewidth]{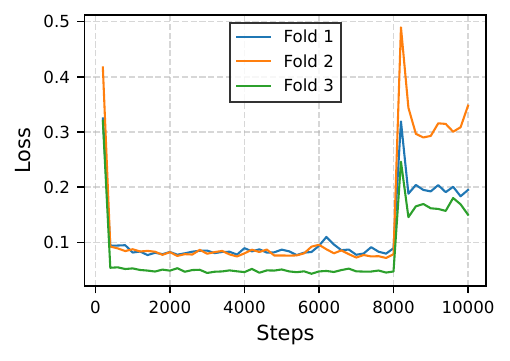}
        \caption{SAC}
    \end{subfigure}

    \caption{Loss Over Steps for Each RL Algorithm}
    \label{fig:loss_rl_all}
\end{figure*}

Figure \ref{fig:loss_rl_all} shows the loss convergence of each algorithm using their best hyperparameter settings. Five of the six algorithms successfully converged over the training period, whereas NFQ diverged. A spike in the loss function is observable across the converging algorithms at exactly the 8k-step mark. DQN shows the highest spike, while CQL shows the least. Following this spike, the loss for these algorithms slightly decreases and stabilizes. This event may correspond to the target synchronization, which was first introduced in \cite{mnih_humanlevel_2015} to stabilize training. Because the target update interval was kept at a default of 8k steps, this created a temporary mismatch between the main and target networks. Conversely, NFQ's divergence may be due to the lack of a separate target network, as it uses the same network for generating the target values and estimating the Q-values, essentially creating a ``moving target" that destabilizes the learning process. Given its influence, careful attention should be paid to the target update interval, as proper tuning could mitigate loss spikes and potentially reduce the loss while yielding a more stable convergence trajectory.


\section{Conclusion}
\label{sec:conclusion}

In this paper, we explored the application of offline RL as a safe, sample-efficient framework for HRI. By bypassing the prohibitive costs and safety risks of online exploration, we developed and evaluated a system architecture that enables a robot to adapt to multimodal human emotions in a board-game scenario. Through our comparative study of offline RL algorithms, we identified BCQ and DDQN as highly robust to hyperparameter variations, whereas CQL proved the most effective at mitigating overestimation bias. Together, these baselines offer a practical, scalable foundation for deploying emotionally intelligent behaviors in real-world applications, such as educational robotics and personal assistants. 

Despite these promising results, several limitations remain. Our constrained offline dataset limits generalizability to demographics and state-action coverage. Future data collection should prioritize diverse interactions and modalities and safely incorporate teleoperation to enrich state exploration. Furthermore, treating HRI as fully observable ignores the latent nature of emotion. Because facial expressions only approximate true internal states, future work should model the environment as a Partially Observable MDP to more accurately track belief distributions over shifting user affect.

\bibliographystyle{IEEEtran}
\bibliography{root}


\end{document}